\documentclass[a4paper]{styles/svproc}  
\usepackage{url}

\usepackage{xspace} 
\usepackage{graphicx}
\usepackage{subcaption}
\usepackage{placeins}
\usepackage[hidelinks]{hyperref}
\usepackage{float}
\usepackage{epsfig}
\usepackage{pifont}
\usepackage{amssymb}
\usepackage{amsmath}
\usepackage[table]{xcolor}
\usepackage{multicol}
\usepackage{enumitem}
\usepackage{cite}
\usepackage{bm}

\newcommand{\pr}[1]{{}}
\newcommand{\todo}[1]{{}}
\newcommand{\com}[1]{{}}
\newcommand{\comb}[1]{{}}
\newcommand{\ouralgo}{HeRO\xspace}
\newcommand{\iros}[1]{{}}

\newcommand{\sbt}{\,\begin{picture}(-1,1)(-1,-3)\circle*{3}\end{picture}\ }

\newcommand\blfootnote[1]{%
  \begingroup
  \renewcommand\thefootnote{}\footnote{#1}%
  \addtocounter{footnote}{-1}%
  \endgroup
}

\begin{document}
\mainmatter              
\title{Towards Resilient Autonomous Navigation of Drones}

\titlerunning{Towards Resilient Autonomous Navigation of Drones}  
%
\author{Angel Santamaria-Navarro* \and Rohan Thakker* \and David D. Fan \and Benjamin Morrell \and Ali-akbar Agha-mohammadi}
%
\authorrunning{A. Santamaria-Navarro et al.} 
%
\tocauthor{Angel Santamaria-Navarro, Rohan Thakker, David Fan, Benjamin Morrell, Ali-akbar Agha-mohammadi}
\institute{
\email{angel.santamaria.navarro@jpl.nasa.gov}
\\ home page:
\texttt{http://angelsantamaria.eu}\\
NASA-Jet Propulsion Laboratory, California Institute of Technology \\ Oak Groove Dr. 4800 Oak Grove Dr, Pasadena, CA 91109, USA,
}

\maketitle              

\begin{abstract}
Robots and particularly drones are especially useful in exploring extreme environments that pose hazards to humans.
To ensure safe operations in these situations, usually perceptually degraded and without good GNSS, it is critical to have a reliable and robust state estimation solution.
The main body of literature in robot state estimation focuses on developing complex algorithms favoring accuracy.
Typically, these approaches rely on a strong underlying assumption: the main estimation engine will not fail during operation.
In contrast, we propose an architecture that pursues robustness in state estimation by considering redundancy and heterogeneity in both sensing and estimation algorithms.
The architecture is designed to expect and detect failures and adapt the behavior of the system to ensure safety.
To this end, we present \ouralgo (Heterogeneous Redundant Odometry): a stack of estimation algorithms running in parallel supervised by a resiliency logic.
This logic carries out three main functions: a) perform confidence tests both in data quality and algorithm health; b) re-initialize those algorithms that might be malfunctioning; c) generate a smooth state estimate by multiplexing the inputs based on their quality.
The state and quality estimates are used by the guidance and control modules to adapt the mobility behaviors of the system.
The validation and utility of the approach are shown with real experiments on a flying robot for the use case of autonomous exploration of subterranean environments, with particular results from the STIX event of the DARPA Subterranean Challenge.
\end{abstract}


\section{Introduction}
\blfootnote{\footnotesize{*Both authors contributed equally to this manuscript. }}
\blfootnote{\footnotesize{The paper has supplementary material including a video showing experimental results of the methods presented in the paper.}}
\blfootnote{\scriptsize{Copyright 2019 California Institute of Technology. U.S. Government sponsorship acknowledged.}}

\pr{Motivate applications and need for resiliency}
Robots are especially useful for accomplishing tasks that would otherwise be impossible or highly dangerous for humans to perform; for instance, search and rescue missions in collapsed zones, package delivery for disaster response or even exploration of other planetary bodies.
As opposed to controlled laboratory environments, deploying autonomous robots in real-world scenarios requires \textit{robustness} and \textit{resiliency} to different types of failures. 
Such failures can include loss of communications, physical damage to sensors or loss of perceptual data in degraded environments (\emph{e.g.}, dust, smoke or fog); and can result in damage to the robot, injuries to humans or failure of mission-critical tasks.
In this work, we focus on a particular class of failures which are related to state estimation.
These failures are often catastrophic since they cascade down through the whole autonomy stack and are of particular importance on aerial robots, where autonomous capabilities are most often limited by the state estimation quality.
\pr{Resiliency Principles}
Hence, we define a set of desirable ''resiliency principles" to reliably operate a system in real-world, challenging environments:  
\begin{itemize}
    \item [\sbt]\textbf{$RP1$: Redundancy} in hardware and algorithms to eliminate single points of failure.
    \item [\sbt]\textbf{$RP2$: Modularity} to easily integrate different components.    
    \item [\sbt]\textbf{$RP3$: Self-recovery} from failures in a distributed component-level fashion.
    \item [\sbt]\textbf{$RP4$: Adaptability} of mobility behaviors based on estimation health status to ensure safety.
    \item [\sbt]\textbf{$RP5$: Real-Time performance} under size, weight and power constraints.
\end{itemize}

\pr{Related work focuses on accuracy and not $RP1$}
These principles are not tackled in most state estimation approaches, which combine data from several sensors into one tightly coupled estimation engine (\emph{e.g.}, \cite{Zhang2015ICRA,Tomic2012AutonUAV,weissmonocular,Santamaria-NavarroAuro18}).
These methods are designed and evaluated for accuracy rather than robustness, possess a single point of failure (no $RP1$) and are not able to recover upon breakdown (no $RP3$), leading to catastrophic failures in the navigation stack.
For example, KITTI dataset \cite{Geiger2012CVPR} benchmarks accuracy of more than 100 approaches that fuse visual information (monocular, stereo or RGB-D cameras), range measurements (SoNAR or LiDAR) and inertial measurements (IMU). 
\pr{Related work partially achieves $RP1$ for sensors}
Methods like \cite{rollinson2013robust,nobili2017heterogeneous,Gosala2018RedundantPA,Carnevaleijars2015} partially consider $RP1$ by exploiting heterogeneity and redundancy of sensors to achieve resiliency for legged, snake-like and race car robots, respectively, but still possess a single point of failure in the software since they use a single estimation engine.
Whereas,~\cite{kim2019resiliency} partially achieves $RP1$ by detecting only measurement failures and searching for a combination of un-compromised sensors.
\pr{Related works assumes estimation never fails $RP4$,$RP3$}
Furthermore, most state-of-art guidance and control modules \cite{morrell2018comparison} either assume that the state estimation module won't fail; or  they try to prevent failures through perception-aware planning \cite{falanga2018pampc} but in their appearance they don't recover from it (no $RP3$) or adapt (no $RP4$).

\pr{Describe \ouralgo and how it achieves resiliency principles}
In this work, we seek to build towards a ``resilient architecture" that follows the resiliency principles stated above, considering the following modules: a) Hardware b) State estimation and c) Guidance and control. 
First, the hardware resiliency can be achieved using mechanical protections, \emph{e.g.}, propeller guards for drones, and using redundant/heterogeneous sensors and actuators ($RP1$).  
Secondly, we can prevent single points of failure ($RP1$) in state estimation by combining heterogeneous redundant odometry algorithms.  
We propose the use of a resiliency logic for supervising the overall estimation pipeline; running confidence checks to verify both sensor data integrity and ``health" of the algorithms; re-initializing the failed sensors or methods ($RP3$); and switching (guaranteeing smooth estimates) among all possible estimation streams to provide the best output.
The resiliency logic also produces a \textit{state quality} measure which is used by the guidance and control module to adapt the mobility behavior of the system ($RP4$), to ensure safety and perform recovery behaviors ($RP3$); for example, switching to velocity control when position estimates are bad or triggering a safe landing behavior using attitude control when velocity estimates are not reliable.
\pr{Modularity and SWaP}
The modular design of \ouralgo ($RP2$) makes it very easy to leverage COTS (commercial-off-the-shelf) odometry solutions.
For example, one can easily combine any number of COTS LiDAR, thermal, and visual odometry algorithms to increase the resiliency of the system.
This modular design also allows for a selection of sensors and algorithms to accommodate for particular computational resources, power or payload, allowing to easily run minimal hardware and software architectures in real-time ($RP5$).

\pr{Structure of rest of the paper}
The remainder of this article is structured as follows.
In the following section, we describe the main ``resiliency architecture", with corresponding concepts and solutions.
Validation and experimental results are presented in Section \ref{sec:validation}, showing the feasibility of the proposed approach through real robot experiments performed live at the STIX event in the DARPA subterranean challenge.
Finally, conclusions are given in Section \ref{sec:conclusions}.

\section{Resiliency Architecture}\label{sec:arch}

\begin{figure}[t]
    \centering
    \includegraphics[width=\linewidth,trim=1cm 0cm 0cm 0cm, clip=true]{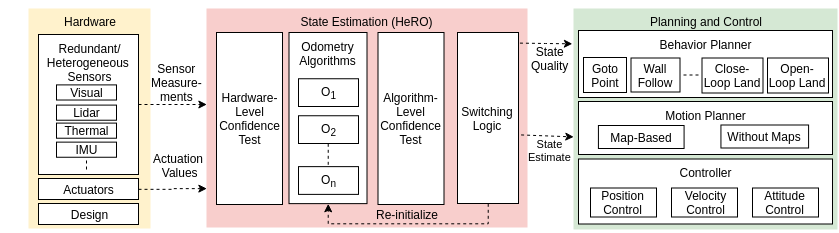}
    \caption{General resiliency architecture.}
    \label{fig:Architecture}
\end{figure}

We propose a framework  designed to autonomously detect and adapt to failures and the key concept behind it is to select the best available state estimation stream, based on confidence checks that assess its quality, and adapt the behavior of the robot by selecting accordingly the appropriate planner and controller.
This general architecture is depicted in Figure~\ref{fig:Architecture} and is described hereafter. 

\subsection{Hardware}

\begin{figure}[t!]
    \centering
    \begin{subfigure}[t]{0.4\textwidth}
        \centering
        \includegraphics[width=\linewidth]{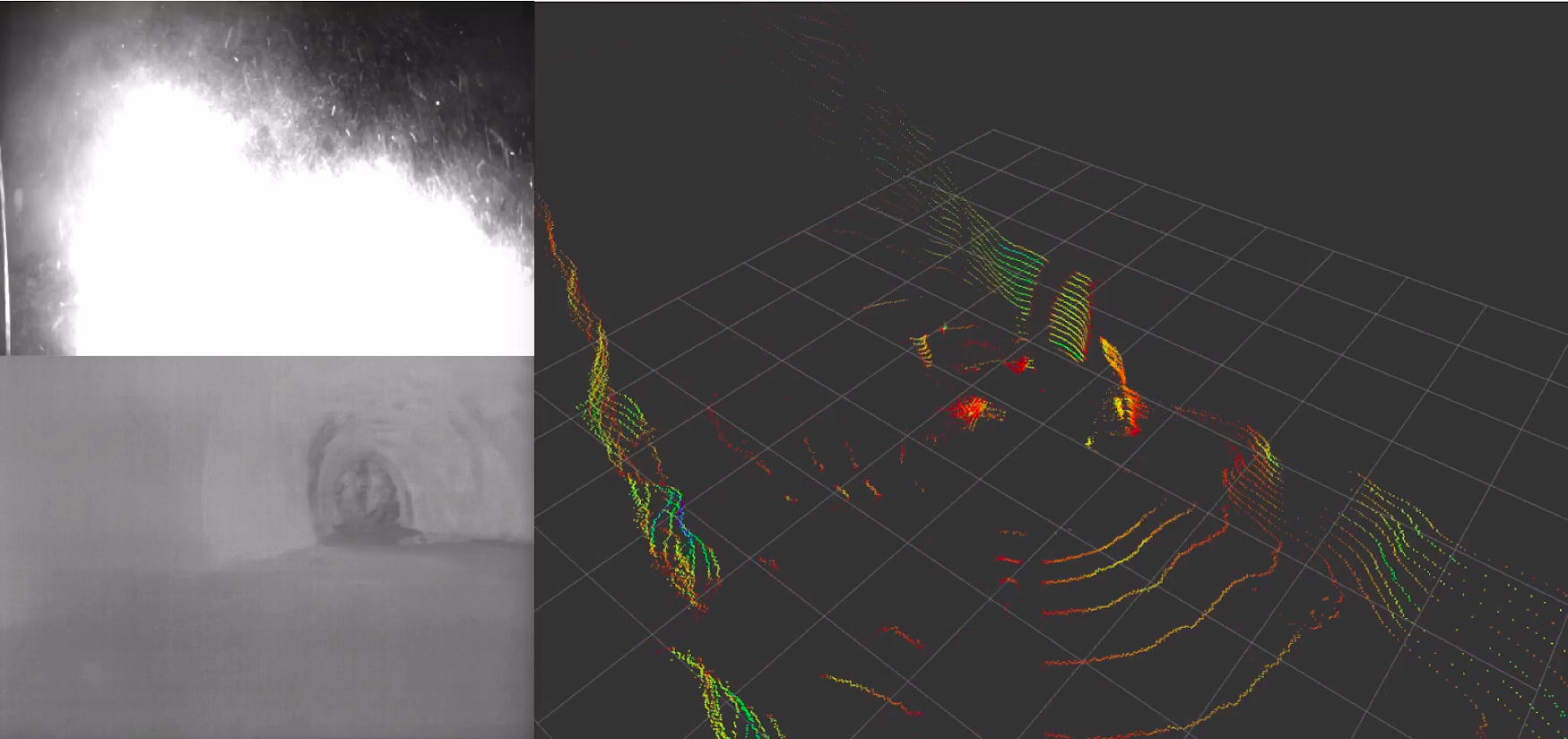}
        \caption{\vspace{1em}}
        \label{fig:visual_bad}
    \end{subfigure}%
    ~ 
    \begin{subfigure}[t]{0.4\textwidth}
        \centering
        \includegraphics[width=\linewidth]{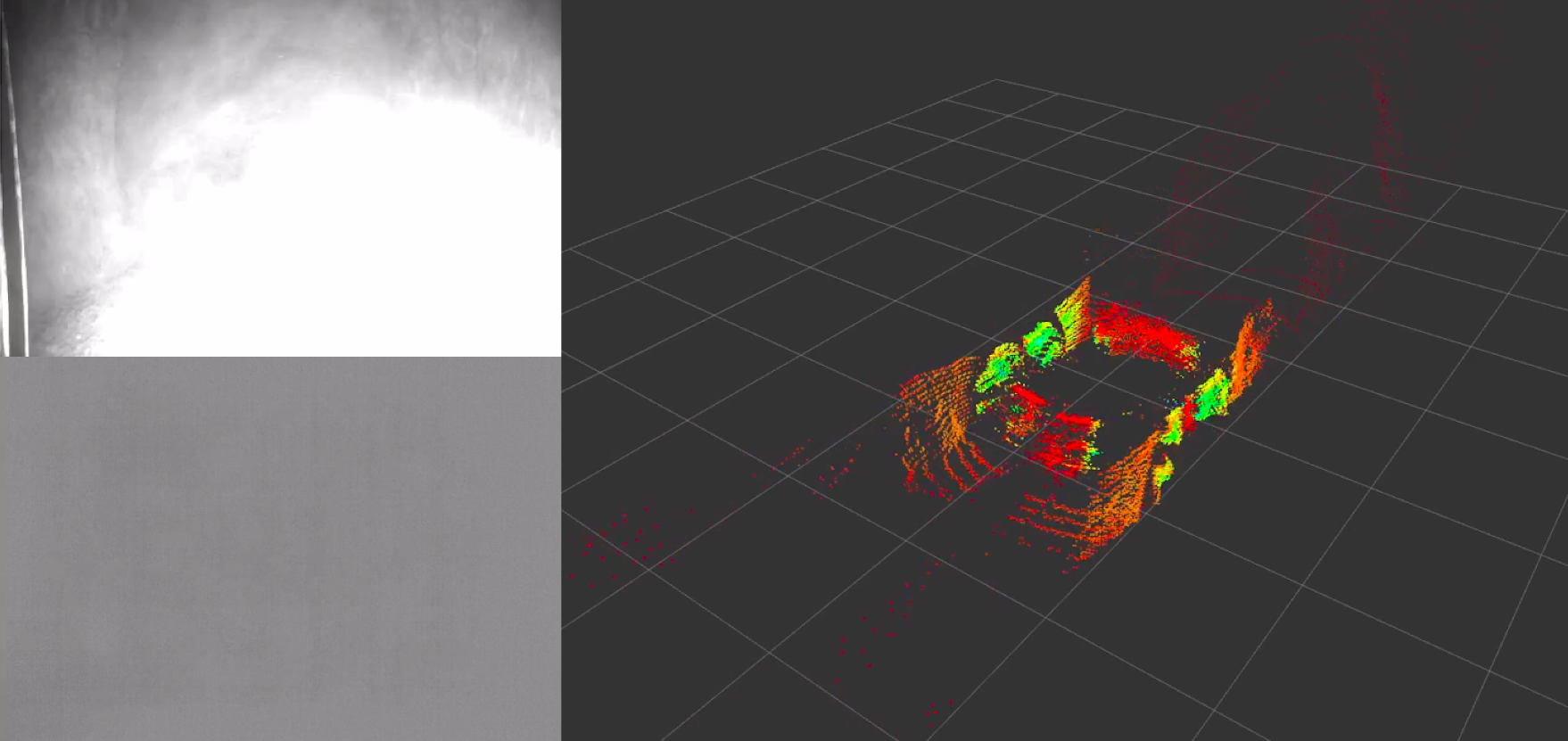}
        \caption{}
        \label{fig:thermal_bad}
    \end{subfigure}
    ~
    \begin{subfigure}[t]{0.4\textwidth}
        \centering
        \includegraphics[width=\linewidth]{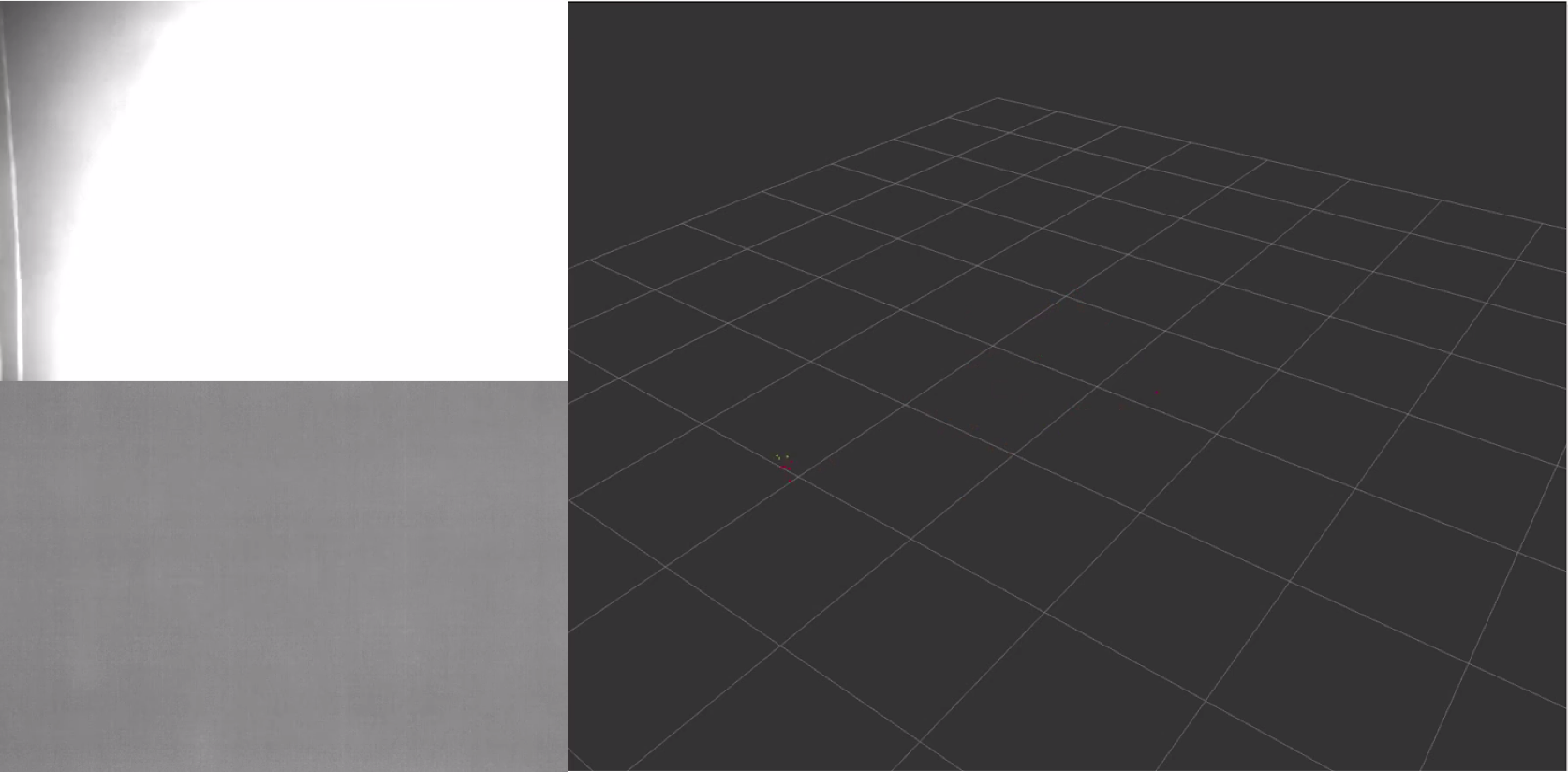}
        \caption{}
        \label{fig:all_bad}
    \end{subfigure}%
    \caption{Example of motivation for heterogeneity and redundancy of methods. For (a)-(c), top left and bottom left are images from visual and thermal cameras, respectively. Right is a point cloud from a 360 degree LiDAR. (a) Visual methods fail with modest dust. (b) Thermal and visual approaches fail with dust and low thermal gradients. (c) LiDAR, thermal and visual fail with intense dust.}
    \label{fig:failing_scenarios}
\end{figure}

The robot hardware consists of a set of heterogeneous and redundant sensors that are used by the state estimation sub-system (\ouralgo).
This can include, but is not limited to, visible\slash infrared\slash thermal cameras, LiDAR (scanners or height sensors), IMU, RaDAR or SoNAR.
The sensor set is recommended to be chosen with a diversity of physical phenomena they are based on, to diversify the possible failure scenarios.
Examples of such failure scenarios are shown in Figure~\ref{fig:failing_scenarios}. 
Here, dust causes issues with visual sensors (Figure~\ref{fig:visual_bad}), as does low thermal gradient for thermal cameras (Figure~\ref{fig:thermal_bad}). 
Intense dust can even cause issues with LiDARs (Figure~\ref{fig:all_bad}), requiring reliance on proprioceptive sensors such as an IMU. 

\begin{figure}[t!]
    \centering
    \begin{subfigure}[t]{0.245\textwidth}
        \centering
        \includegraphics[width=\linewidth, trim=1cm 9.5cm 12cm 1cm, clip=true]{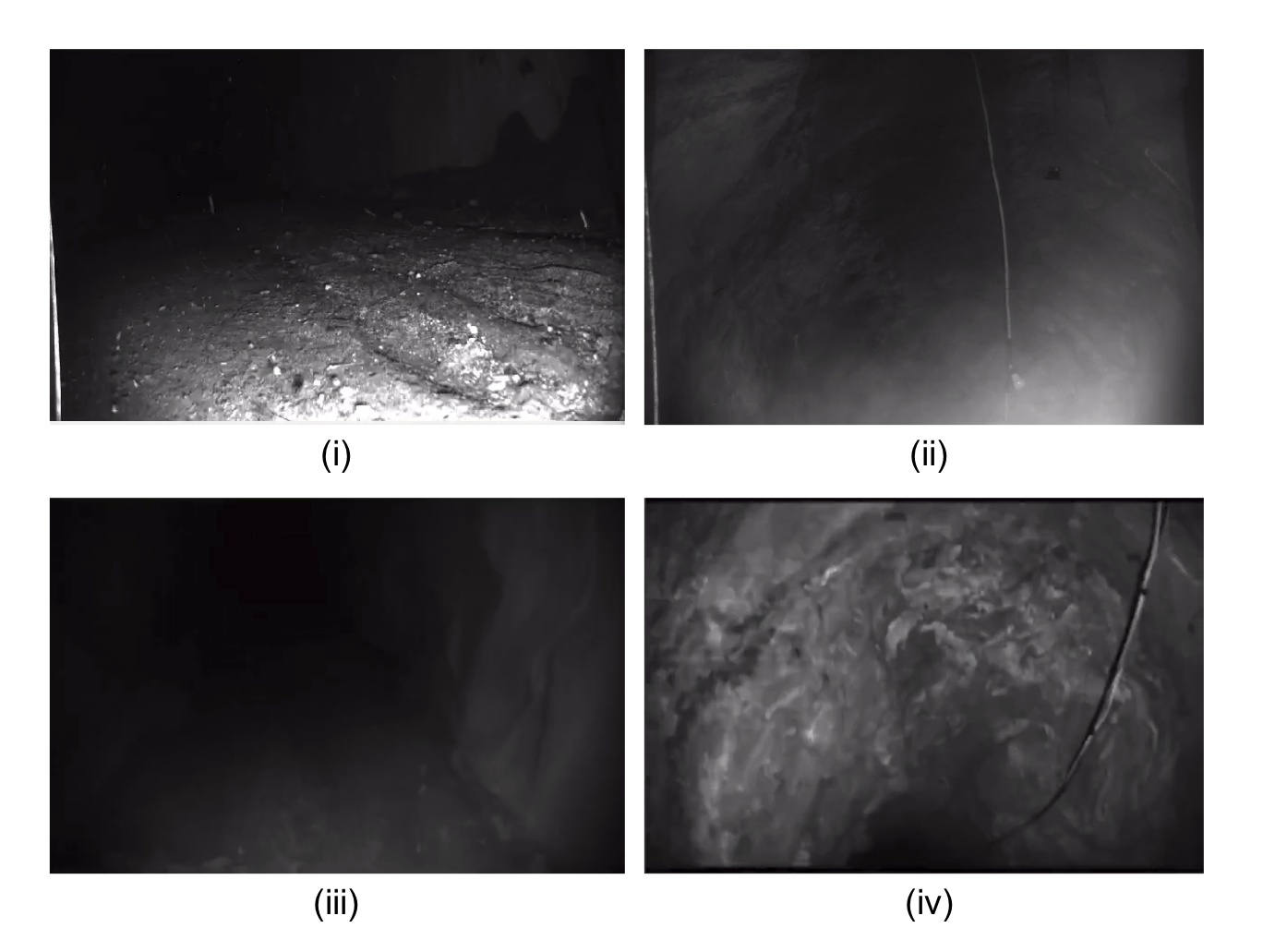}
        \caption{}
        \label{fig:forwardcam}
    \end{subfigure}
    \begin{subfigure}[t]{0.245\textwidth}
        \centering
        \includegraphics[width=\linewidth, trim=12cm 9.5cm 1cm 1cm, clip=true]{fig/DifferentDirections.png}
        \caption{}
        \label{fig:upwardcam}
    \end{subfigure}%
    \begin{subfigure}[t]{0.245\textwidth}
        \centering
        \includegraphics[width=\linewidth, trim=1cm 1.3cm 11.9cm 9cm, clip=true]{fig/DifferentDirections.png}
        \caption{}
        \label{fig:forwardcam_tunnel}
    \end{subfigure}
    \begin{subfigure}[t]{0.245\textwidth}
        \centering
        \includegraphics[width=\linewidth, trim=12cm 1.3cm 1.5cm 9.5cm, clip=true]{fig/DifferentDirections.png}
        \caption{}
        \label{fig:upwardcam_tunnel}
    \end{subfigure}
    \caption{Contrasts in visual data from cameras pointing to different directions for different robot positions. Here, to overcome the dark environment, we use on-board illumination. (a) Forward facing near the ground: good lighting and texture. (b) Upward facing near the ground: poor lighting and limited detail. (c) Forward facing when flying in a large tunnel: poor illumination and no texture. (d) Upward facing in the same scenario as (d): good illumination and ceiling details.}
    \label{fig:visual_foward_up}
\end{figure}

In addition to hardware considerations, an important factor to reduce the chances of failure is with correct sensor placements. 
For example, a VIO running on a drone has higher chances of failure if using a forward facing camera compared to a downward facing camera while flying at low altitudes.
However, in the latter case, the VIO is susceptible to failure if the drone is tilting or moving very close to the ground at high speeds; or if there is a lack of visual features on the ground. 
Additionally, there may be different amounts of visual texture or lighting in the different directions. 
This is the case in dark underground environments, where illumination of the scene comes from light sources on the robot (see Figure~\ref{fig:visual_foward_up}). 
Hence, using multiple camera orientations reduces the chances of overall system failure at the cost of more mass, power and computational resources. 

Finally, a robust mechanical design of the robot hardware can allow autonomy to have significantly higher tolerances to avoid failure. 
Notice how incorporating redundancy, heterogeneity and mechanical robustness on robots such as drones, can result in significantly lower flight time, due to increased weights of sensors, however it can highly increase the probability of mission success.

\subsection{Heterogeneous-redundant odometry  (\ouralgo)}
Our objective is to estimate the following robot states with respect to a ``world" frame along with their quality:\\
\begin{tabular}{cll}
&\hspace{1cm}$\vec{p} \in \mathbb{R}^3$ &\hspace{0.5cm}Position represented in world frame\\  
&\hspace{1cm}$\vec{R} \in SO(3)$ & \hspace{0.5cm}Orientation represented in world frame\\
&\hspace{1cm}$\vec{v} \in \mathbb{R}^3$ & \hspace{0.5cm}Linear velocity\\
&\hspace{1cm}$\vec{\omega} \in \mathbb{R}^3$ & \hspace{0.5cm}Angular velocity \\
&\hspace{1cm}$\vec{a} \in \mathbb{R}^3$ & \hspace{0.5cm}Linear acceleration\\
&\hspace{1cm}$\vec{\alpha} \in \mathbb{R}^3$ & \hspace{0.5cm}Angular acceleration\\
&\hspace{1cm}$Q_i \in \{Good, Bad\}$ & \hspace{0.5cm}Quality of $i$, with $i \in [\vec{p}, \vec{R}, \vec{v}, \vec{\omega}, \vec{a}, \vec{\alpha}]$
\end{tabular}\\
These states are used (when available) for motion planning and control, whereas the quality metrics are used by the behavior planner to select the appropriate mobility service for the current mission task.
Note that all attributes are represented in robot body frame unless stated otherwise. 
Moreover, we restricted the quality of the state to binary values although it can be easily generalized to higher resolutions and even continuous representations.

To obtain the best available estimation, we use \ouralgo, consisting of three main components: a) A stack of odometry algorithms running in parallel, b) A set of confidence checks on the quality of the estimation and c) A resiliency logic.
All these modules are detailed in the following.

\subsubsection{\bf Odometry Algorithms}
\hfill\\
Any single estimation algorithm will have circumstances in which may fail; circumstances that are particular to the sensor/algorithm combination.
Hence, we run in parallel a stack of heterogeneous odometry algorithms (\emph{e.g.}, visual, LiDAR, thermal, RaDAR or SoNAR), to increase the probability of overall success by having non-overlapping failure scenarios. 
There is no special requirement on the type of algorithm, with the ability to incorporate either tightly or loosely coupled approaches. 
However, to take advantage of all possible mobility services, there is a need for estimating position, orientation, velocity and, ideally, acceleration.
If the method is only estimating positions, it can be fused with IMU measurements (in a loosely coupled fashion) to obtain the rest of the state. 
This might be the case for pure visual or LiDAR odometry methods (VO and LO, respectively).

\subsubsection{\bf Confidence Checks and Health Monitoring}
\hfill\\
The detection of malfunctioning odometry streams is a fundamental feature to select the best state estimation. 
Hence, it is of interest to define the possible failures in this context.
At a high level, a failure is any estimation that jeopardizes robot controls or planning.
Early identification of failures can allow for system recovery or, if recovery is not possible, performing safety fall-backs to keep the robot integrity.
\ouralgo detects the failures by running confidence checks at different levels of implementation.
First, \ouralgo performs confidence checks at hardware-level by checking the data from sensors. 
As an example, this type of checks in case of using cameras or LiDARs might include (but are not limited to):
\begin{itemize}
    \item[\sbt] Rate of sensor output. 
    \item[\sbt] Overall image intensity or its variation within the image.
    \item[\sbt] Distance between first and last return of a LiDAR beam.
    \item[\sbt] Number of invalid scan points.
\end{itemize}
\begin{figure}[t]
    \centering
    \includegraphics[width=0.6\linewidth]{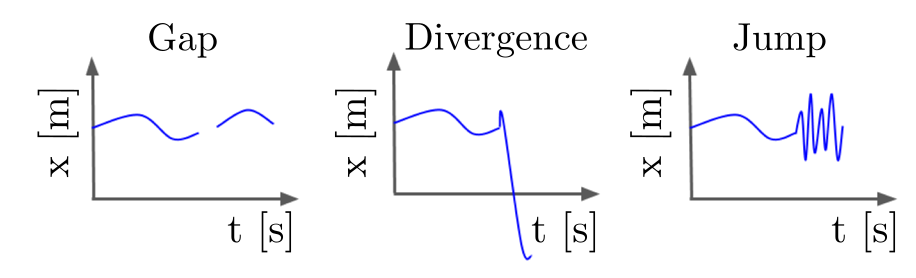}
    \caption{Different failure modes for odometry algorithms.}
    \label{fig:failure_modes}
\end{figure}
Next, \ouralgo performs confidence checks using data from the odometry algorithms, where the goal is to catch the failures depicted in Figure~\ref{fig:failure_modes}, ideally before they occur or at an early stage. 
These failures can come in a variety of forms, being a gap in the state updates, a divergence of the estimate, or rapid jumps. 
The failure could be caused by limitations of the sensor or by the overall odometry algorithm. 
The checks to detect these failures include:
\begin{itemize}
    \item[\sbt] Rate of algorithm output (to catch gaps).
    \item[\sbt] Rate of change of the position/velocity estimate (to catch jumps).
    \item[\sbt] Trace of the estimation covariance matrix (to catch divergence).
\end{itemize}
Notice how we can also run checks dependant on particularities of the methods; for instance number or quality of features tracked; or number or quality of inlier features (\emph{e.g.}, from matching/RANSAC).

\clearpage
\subsubsection{\bf Resiliency Logic}
\hfill\\

The resiliency logic in \ouralgo is based on the various odometry estimates and respective confidence checks, to output a resulting state with its quality measure, which will be used by the planner and controller.
Moreover, it supervises the individual odometry estimates and requires them to re-initialize in case of detecting failures.

An initial version of the logic can use a predefined ranking of odometry algorithms to select the best estimation method, thus selecting the approach with highest priority that is not failing. 
When that algorithm fails a confidence check, it is required to re-initialize while the next algorithm in the ranking is selected. 
These switches are provoked only in case of failure of the current source in order to minimizes the number of switches. 
A more advanced logic can consider a selection criteria weighted by the quality of the algorithm.
This comparison between streams also provides a third avenue for confidence checks; for instance, with three or more streams a voting scheme can be used to identify outliers.

The resiliency logic has to guarantee continuity in the resulting estimate while switching between streams. 
This consistency in the output can be achieved by iteratively composing an incremental state from the selected odometry source to the state from the previous estimate (\emph{e.g.}, a method estimating pose will re-start with its origin at {\bf 0}, hence if we switch to it after its re-initialization, we have to concatenate its estimation with the previous resulting state of \ouralgo).


The resulting quality of the estimation is generated depending on the confidence checks in the odometry source being used. 
Depending on this quality we can enable different mobility services, shown in Table~\ref{tab:state_quality} and detailed in the following section.
Note that in Table~\ref{tab:state_quality} we incorporated the platform height estimation ($^g{p}_z$) as a separate column because it can be directly used for the controller (altitude hold) and is an intermediate step between having all position estimates (${\bf p}$) or just velocities (${\bf v}$). Similarly, this sensor can be used to estimate vertical velocities ($v_z$). 
These velocity-only states can appear either because they are directly provided by a method (\emph{e.g.}, optical flow) or because the logic decided to keep using an approach which is partially failing; for instance, using the output of a method that runs on IMU only while we reset the other update sources. 
In the latter, velocities can be trusted for short periods.
In Table~\ref{tab:state_quality}, we assume the IMU is not failing, providing good attitude (${\bf R}$), angular velocities (${\bm \omega}$) and accelerations (${\bf a}$,${\bm \alpha}$) with biases estimated by those methods previously working. 
If that is not the case we consider an immediate landing in ``open loop" control mode.

\begin{table}[t]
    \centering
    \begin{tabular}{|c|c|c|c|c|c|c|}
    \hline
    Case & \multicolumn{5}{c|}{State Quality} & Mobility \\
    \hline
    No. & ${\bf p}$ & $^gp_z$ & $v_x, v_y$ & $v_z$ & ${\bf R}$, ${\bm \omega}$, ${\bf a}$, ${\bm \alpha}$ & \\
    \hline
    1 & \cellcolor{green!25} & \cellcolor{green!25} & \cellcolor{green!25} 
      & \cellcolor{green!25} & \cellcolor{green!25} & Global\\
    \hline
    2 & \cellcolor{red!25} & \cellcolor{green!25} & \cellcolor{green!25} 
      & \cellcolor{green!25} & \cellcolor{green!25} & Local\\
    \hline
    3 & \cellcolor{red!25} & \cellcolor{red!25} & \cellcolor{green!25} 
      & \cellcolor{green!25} & \cellcolor{green!25} & Local\\
    \hline
    4 & \cellcolor{red!25} & \cellcolor{green!25} & \cellcolor{red!25} 
      & \cellcolor{green!25} & \cellcolor{green!25} & Closed Loop on $z$\\
    \hline    
    5 & \cellcolor{red!25} & \cellcolor{red!25} & \cellcolor{red!25} 
      & \cellcolor{red!25} & \cellcolor{green!25} & Attitude\\
    \hline
    \end{tabular}
    \caption{Summary of available mobility services given different estimation qualities resulting from \ouralgo. The colour code is \colorbox{green!25}{Good} \colorbox{red!25}{Bad}.}
    \label{tab:state_quality}
\end{table}

\subsection{Planning and Control}
The estimated state will be used for planning and control (see Figure~\ref{fig:Architecture}). 
In this work we consider three main layers for these modules:
\begin{itemize}
    \item[\sbt]{\bf Behavior planning:} A state machine which chooses the type of behaviors the robot should execute depending on the available quality of the estimated state. The possible decisions are depicted in the right-most column of Table~\ref{tab:state_quality}.
    \item[\sbt]{\bf Motion planning:} Generates desired trajectories for the robot according to the decisions of the behavior planner. 
    \item[\sbt]{\bf Control:} Tracks desired trajectories and closes the loop directly on the provided state estimates from \ouralgo.
\end{itemize}
With regards to the state estimation quality, we distinguish the following mobility services, which include the above mentioned layers:
\begin{itemize}
\item[\sbt]{\bf Global:} 
This group of services require good robot position estimates and may include (but are not limited to) building occupancy-grid maps, running path planning algorithms for reaching a desired location in the map, or reasoning and executing high-level \textit{global} goals.
Notice how many of these planning and control algorithms rely on continuous estimates of position, which \ouralgo provides.
\item[\sbt]{\bf Local:}
Without reliable position estimates, the robot is restricted to plan and execute \textit{local} actions depending on reliable attitude, velocity and acceleration estimates. Examples of such behaviors can include wall-following, obstacle avoidance or hover-in-place. 
These behaviors are often robust and locally optimal\cite{MaplessPlanner}; however, they often require some tuning and assumptions on the topology of the environment (\emph{e.g.}, flying through a long tunnel can be accomplished with wall-following behavior). 
\item[\sbt]{\bf Attitude:} If it is not possible to maintain high-quality estimation of either positions or velocities, then specific \textit{attitude} behaviors can be performed to keep the integrity of the robot or even to recover.
For example, a drone can land using attitude control and slowly reduce its thrust. Once landed, the chances of being able to re-initialize odometry algorithms dramatically increase (\emph{e.g.}, camera artifacts like motion blur are reduced by not moving or the effect of dust is mitigated by not spinning the propellers). 
\end{itemize}
Notice how some odometry algorithms can give mixed results on their state estimation quality; for instance, estimating reliably velocities but intermittently the positions. 
In this work, we take advantage of as much reliable state information as possible.
An example is obtaining height estimations from a single ranger, helping to close the loop only on the vertical axis (see case 4 in Table~\ref{tab:state_quality}).
\section{Validation and experiments}\label{sec:validation}
In this section, we show the validity of our state estimation framework with a specific implementation.
We set up the resiliency logic supervising three odometry algorithms using vision, a LiDAR and an IMU as sensing modalities. The first algorithm is a loosely coupled VIO with stereo cameras facing forward, a tightly coupled VIO with a monocular camera facing up, and a LiDAR-inertial odometry (LIO) algorithm with a 360 degree, 16 channel LiDAR. This setup provides redundancy of sensors and heterogeneity in the algorithmic solution. 
In the following we describe in detail the hardware, odometry algorithms, confidence checks and multiplexing approaches used.
Moreover we include experimental results obtained from the STIX event of the DARPA subterranean challenge\footnote{https://www.subtchallenge.com}. 

\subsection{Hardware}
\begin{figure}[t]
    \centering
       \includegraphics[width=0.6\textwidth]{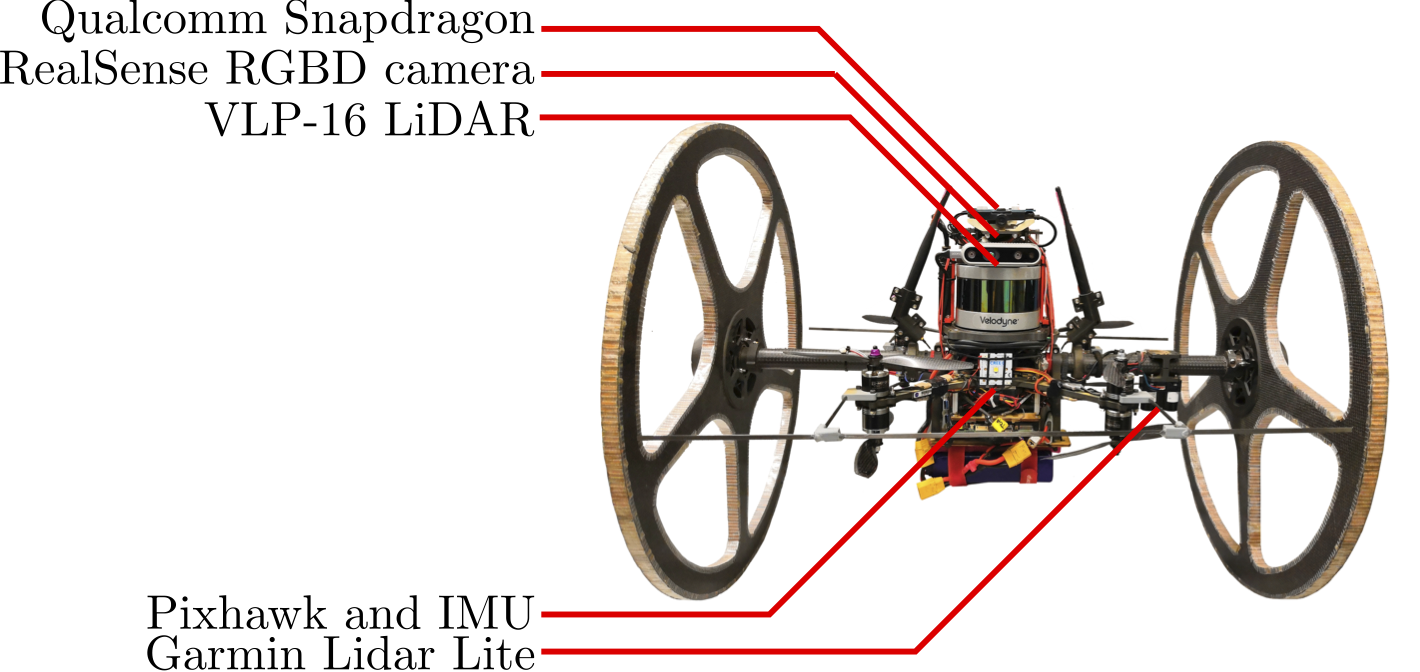}
      \caption{The \emph{Roll-o-copter}: a hybrid terrestrial and aerial multi-rotor robot}
      \label{fig:rollo}
\end{figure}

We make use of the \emph{Roll-o-copter}: a hybrid terrestrial and aerial multi-rotor equipped with two passive wheels (see Figure~\ref{fig:rollo}). 
Although this robot is designed to take advantage of both aerial and ground terrains, in this work we focus on the aerial mobility of the vehicle and use it as flying-only platform.  
In flight, \emph{Roll-o-copter} behaves like a normal multi-rotor, with a standard set of electronic speed controllers (ESCs), motors and propellers. 
It possesses a Pixhawk\footnote{http://www.pixhawk.org} v2.1 as flight controller, as well as an on-board Intel NUC i7 Core computer. 
In addition to the standard multi-rotor hardware, we equipped the robot with the following sensors:

\begin{itemize}
\item[\sbt] {\bf RealSense\footnote{https://software.intel.com/en-us/realsense/d400} RGBD camera.} Composed by a stereo pair of infrared (IR) cameras, an RGB camera and a structured light IR projector. 
\item[\sbt] {\bf Velodyne $360^\circ$ VLP-16\footnote{https://velodyneLiDAR.com/vlp-16.html} LiDAR.} Rotary head with 16 LiDAR rangers, providing a point cloud with a $360^\circ$ azimuth angle and $\pm15^\circ$ elevation angle field of view.
\item[\sbt] {\bf Garmin LiDAR Lite v3.} We use a 1D LiDAR pointing downwards.
\item[\sbt] {\bf Pixhawk v2.1 flight controller.} Composed by an on-board IMU (3-axis gyroscopes and accelerometers) and processing, including a Kalman filter for its own state estimation.
\item[\sbt] {\bf Qualcomm Snapdragon\footnote{https://developer.qualcomm.com/hardware/qualcomm-flight-pro}.} A self-contained flight controller accompanied by an IMU (3-axis gyroscopes and accelerometers), a high resolution forward facing camera and a low resolution upward facing camera for tightly coupled VIO.
\end{itemize}

\subsection{Algorithm Stack}

We validate the \ouralgo approach with two VIOs and one LIO in the stack of methods, which provide the required capabilities to fly in complex and perception-challenging environments. 
These methods are running in parallel at different frequencies and using different sensor sources ($RP1$), as detailed in the following.

\begin{itemize}

\item[\sbt] {\bf Infrared Stereo Visual Inertial Odometry.}
We take advantage of ORB-SLAM2 (OS2) \cite{orbslam2mur2017} running using images from the IR stereo camera (RealSense RGBD). 
This approach produces 6D pose estimates (3D translations and 3D rotations) and can run up to 60Hz.
These estimations are fused with IMU data (running at a frequency of 1kHz) using an EKF (running in the Pixhawk flight controller).
Using a stereo odometry algorithm enables us to re-initialize it in flight without the dependence of parallax movements such as in monocular VIO methods.
To have continuous odometry estimates when available we disable the loop closure modules of this approach.

\item[\sbt] {\bf Monocular Visual Inertial Odometry.}
We use the Snapdragon Flight platform from Qualcomm (QSF) running their COTS VIO from mvSDK as the second source of odometry.
As in the case of OS2 odometry, here we also incorporated some modifications to allow re-initialization during flight ($RP3$).
The incorporation of this approach also demonstrates the capability of our framework to easily integrate closed-source commercial VIO solutions ($RP2$).
The state estimation runs at 25Hz.

\item[\sbt] {\bf LiDAR Inertial Odometry.}
The third source of odometry consists on fusing 6D pose estimates from a LiDAR odometry (LO) approach with IMU data within a regular EKF scheme. 
The pose estimates from the LO are produced at 20Hz and we use the same IMU as for the IR Stereo VIO, but this time with the data externalized from the Pixhawk flight controller running at 200Hz.
\end{itemize}
Notice how the above choices allow us to use three different sensing modalities (IR stereo forward-facing, monochrome monocular upward-facing and LiDAR) with three different estimation algorithms ($RP1$). 
All sensors and approaches are prone to different possible failures, minimizing the occasions where they all fail simultaneously, thus achieving the required level of robustness, as show in the following sections and the accompanying video.

\subsection{Resiliency logic}

In these experiments we use the following confidence tests, set by observing the types of failures the methods are prone to.

\begin{itemize}
\item[\sbt] {\bf Frequency:} The most common mode of failure of OS2 occurs due to feature tracking failures when the drone is executing a fast motion or due to presence of featureless environments. 
In this case, the frequency of the measurement updates goes down while failing. This policy helps catching ``data gaps".
\item[\sbt] {\bf Estimation covariance:} The uncertainty of the estimated pose from QSF significantly increases during failure, the state estimate starts to diverge. 
Hence, we detect these failures by setting a threshold on the trace of the estimation covariance matrix (experimentally determined). This policy helps catching ``data divergence".
\item[\sbt] {\bf Sudden position changes:} If the estimation method results are inconsistent, it might still produce an output although the covariance of the estimations might not reflect it. To detect these failures we set a confidence check on sudden position changes to catch ``data jumps". 
\end{itemize}

\subsection{Navigation in a mine, example of usage in the STIX event of DARPA subterranean Challenge}

\begin{figure}[t]
    \centering
    \begin{subfigure}[t]{0.325\textwidth}
        \centering
        \includegraphics[width=\linewidth, trim=0cm 0cm 0cm 0cm, clip=true]{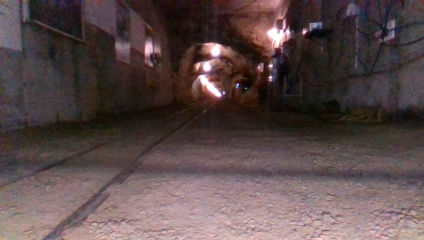}
        \caption{}
        \label{fig:miami_rollo_picture_a}
    \end{subfigure}
    \begin{subfigure}[t]{0.325\textwidth}
        \centering
        \includegraphics[width=\linewidth, trim=0cm 0cm 0cm 0cm, clip=true]{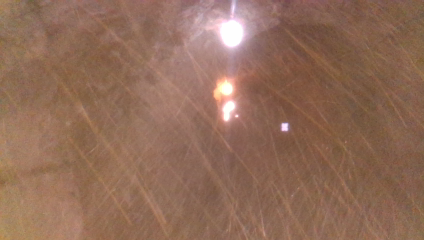}
        \caption{}
        \label{fig:miami_rollo_picture_b}
    \end{subfigure}
    \begin{subfigure}[t]{0.325\textwidth}
        \centering
        \includegraphics[width=\linewidth, trim=0cm 2.38cm 0cm 0cm, clip=true]{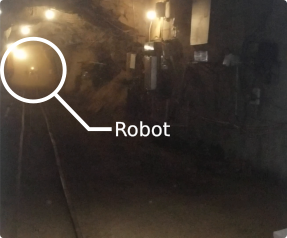}
        \caption{}
        \label{fig:miami_rollo_picture_c}
    \end{subfigure}
    \caption{Real experiment. (a) Frontal camera with the robot idle on the ground. (b) Frontal camera during flight. (c) External view during flight.}
    \label{fig:miami_rollo_picture}
\end{figure}

The experimental scenario is part of the DARPA subterranean challenge\footnote{https://www.subtchallenge.com/} and, specifically, we show results from the official integration event (STIX). 
We provide an accompanying video with some of the flights.
In these experiments we deploy the \emph{Roll-o-copter} in the entrance of an iron ore mine and set an autonomous exploration mission. 
Figure \ref{fig:miami_rollo_picture} shows an example of one of the experimental runs, demonstrating how challenging the environment is in terms of perception and justifies the use of a resilient state estimation approach (\emph{e.g.}, notice the dusty environment comparing Figure~\ref{fig:miami_rollo_picture_a} where the robot did not start with~\ref{fig:miami_rollo_picture_b} and~\ref{fig:miami_rollo_picture_c} while flying).
In these experimental runs of the main STIX event we were able to accomplish our missions with \emph{Roll-o-copter}, exploring two different entrances of the mine and validate our resilient state estimation strategy.

\begin{figure}[h!]
  \centering
  \begin{tabular}{c}
    \begin{subfigure}[t]{0.85\textwidth}
        \centering
        \includegraphics[width=\linewidth,trim=0cm 0.95cm 0.5cm 0.2cm, clip=true]{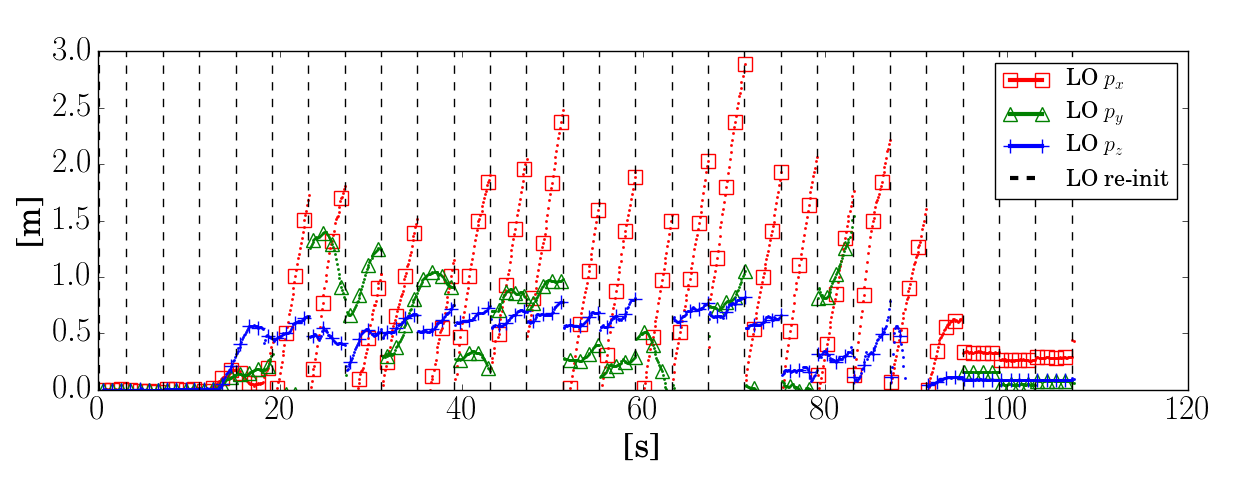}
        \caption{LiDAR odometry (LO) X-Y-Z positions.}
        \label{fig:miami_rollo_data_a}
    \end{subfigure}%
    \\
    \\
    \begin{subfigure}[t]{0.85\textwidth}
        \centering
        \includegraphics[width=\linewidth,trim=0cm 0.95cm 0.5cm 0.2cm, clip=true]{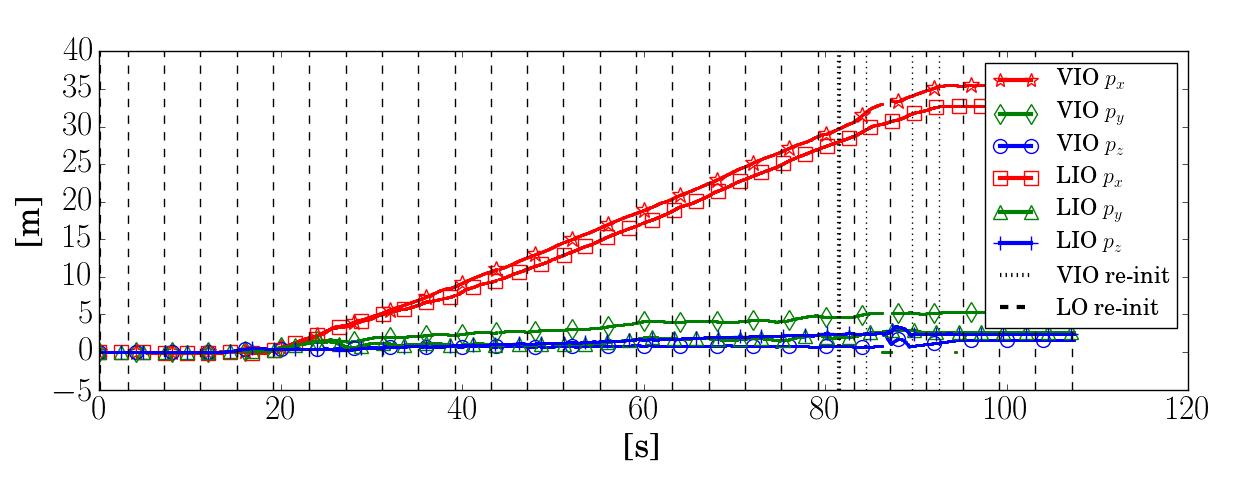}
        \caption{VIO (OS2) and LIO X-Y-Z position estimates.}
        \label{fig:miami_rollo_data_b}
    \end{subfigure}%
    \\
    \\
    \begin{subfigure}[t]{0.85\textwidth}
        \centering
        \includegraphics[width=\linewidth,trim=0cm 0.95cm 0.5cm 0.2cm, clip=true]{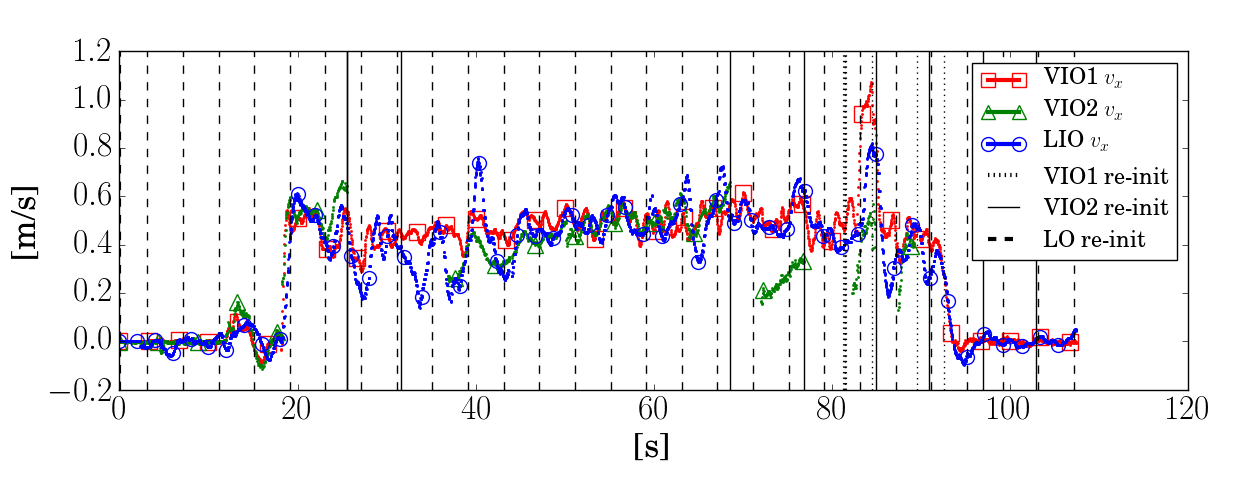}
        \caption{VIO (OS2) and LIO X velocity estimates.}
        \label{fig:miami_rollo_data_c}
    \end{subfigure}%
  \end{tabular}  
    \caption{Results from a real experiment with the \emph{Roll-o-copter} flying autonomously in a tunnel (see Figure~\ref{fig:miami_rollo_picture}).}
    \label{fig:miami_rollo_data}
\end{figure}

An example of these runs (mine entrance 1) is shown in Figure~\ref{fig:miami_rollo_data} with regular re-initialization of LIO. 
Here we show the different types of re-initialization, depending on the effected odometry stream. 
A first case consists of partly re-initializing a loosely-coupled odometry approach (LO, in this case), for example resetting one of the inputs of the IMU fusion algorithm. 
In these cases, we can detect discontinuities by checking the individual estimations (\emph{e.g.}, LO). 
To avoid inconsistencies, whenever we re-initialize a particular singular source (\emph{e.g.}, VO or LO) that is subsequently fused with an IMU, we keep track of pose discontinuities with a transform manager, similarly as done by the resiliency logic for its estimation output while switching between odometry streams. 
Figure~\ref{fig:miami_rollo_data_a} shows this behavior, and how re-initializing manually every 5s a LiDAR odometry (LO) provokes LO position estimates to jump. 
Thanks to the transform manager, the fusion of this LO with an IMU (LIO) is kept consistent and continuous (see Figure~\ref{fig:miami_rollo_data_b} where we show the LIO output and an OS2 VIO for comparison purposes).
Figure~\ref{fig:miami_rollo_data_c} shows an example of resulting x-body axis velocities of all methods in the \ouralgo stack, running two VIOs (VIO1 is OS2 and VIO2 is QSF) and the LIO test for the same experiment.
In this run the resiliency logic was directly selecting VIO1 (used for autonomous navigation) and it requested several times a re-initialization of the VIO2 due to failures.

\begin{figure}[h!]
\centering
\vspace{-1.2em}
\begin{tabular}{cc}
  \rotatebox{90}{\hspace{1.5cm}$p_x$} & \includegraphics[width=0.7\linewidth,trim=0.4cm 2cm 0cm 0.9cm, clip=true]{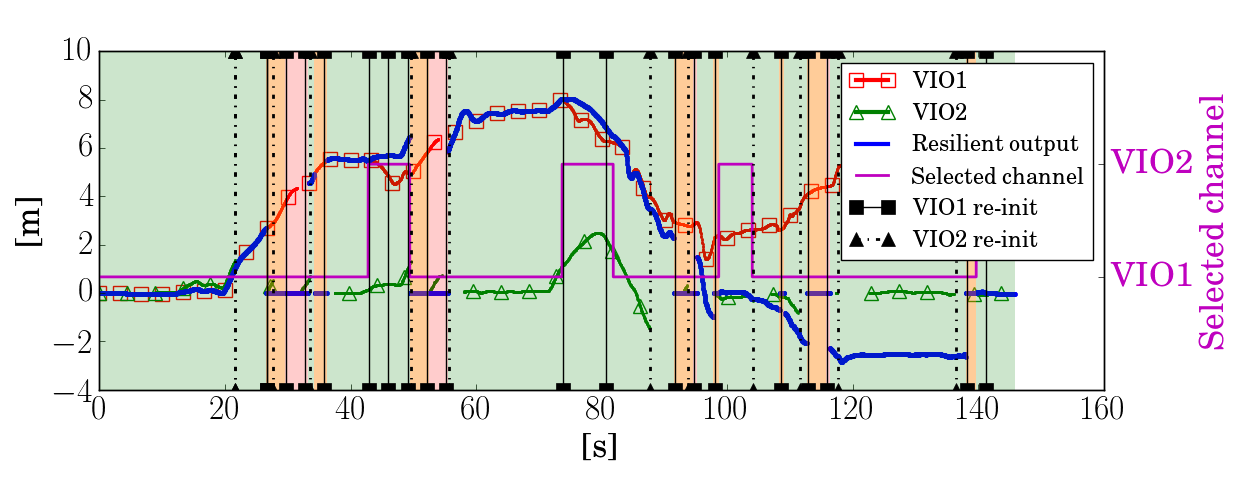} \\
  \rotatebox{90}{\hspace{1.5cm}$p_y$} & \includegraphics[width=0.7\linewidth,trim=0.4cm 2cm 0cm 0.9cm, clip=true]{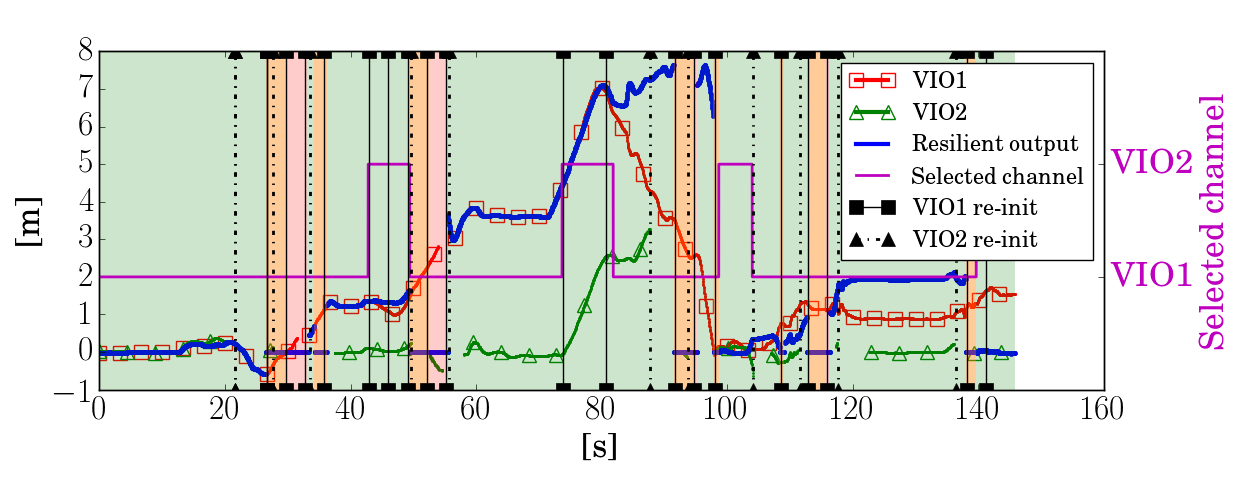} \\
  \rotatebox{90}{\hspace{1.5cm}$p_z$} & \includegraphics[width=0.7\linewidth,trim=0.4cm 2cm 0cm 0.9cm, clip=true]{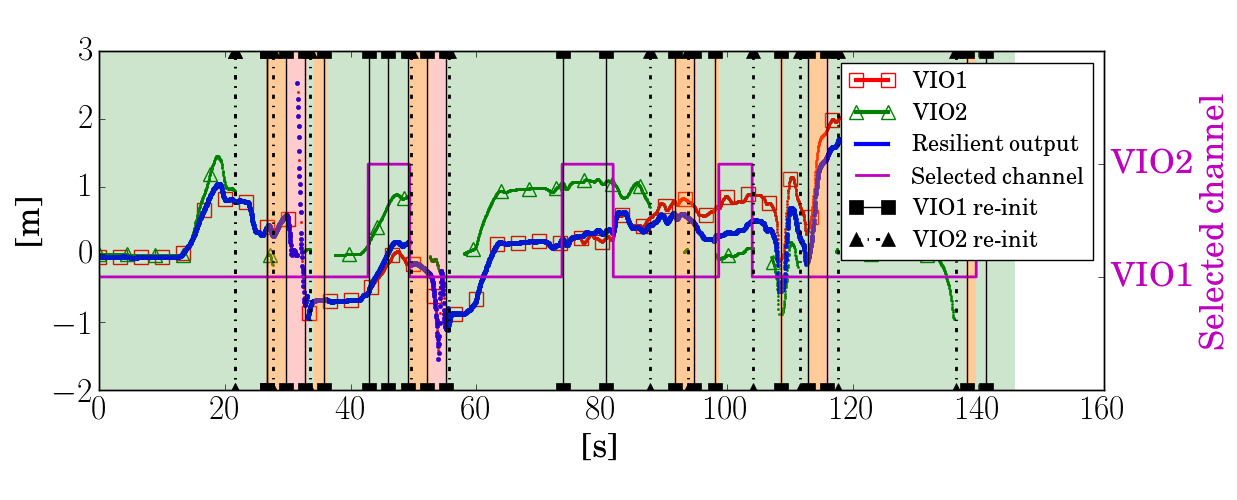} \\
  \rotatebox{90}{\hspace{1.5cm}$v_x$} & \includegraphics[width=0.7\linewidth,trim=0.4cm 2cm 0cm 0.9cm, clip=true]{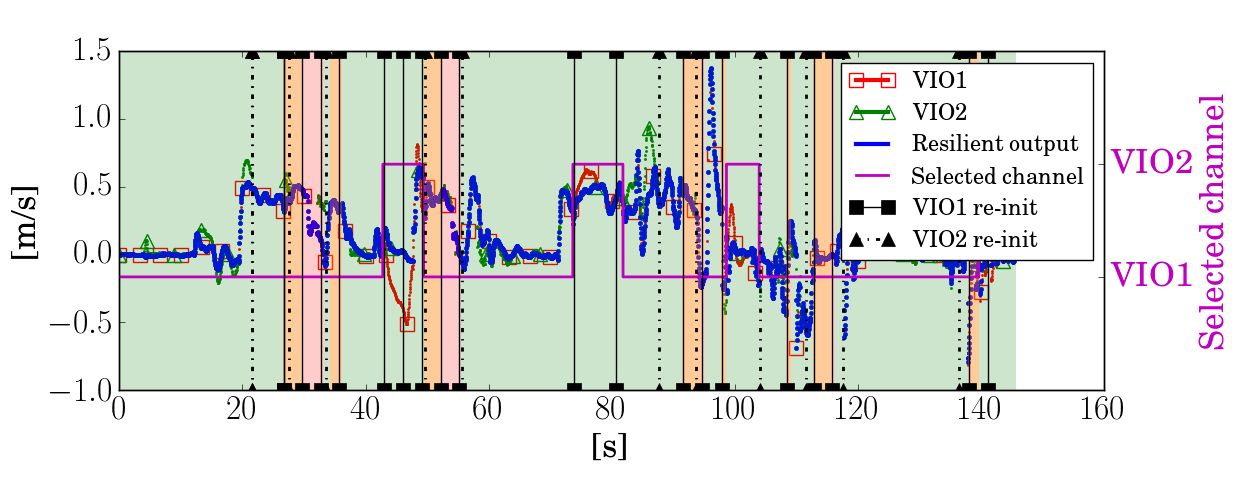} \\
  \rotatebox{90}{\hspace{1.5cm}$v_y$} & \includegraphics[width=0.7\linewidth,trim=0.4cm 2cm 0cm 0.9cm, clip=true]{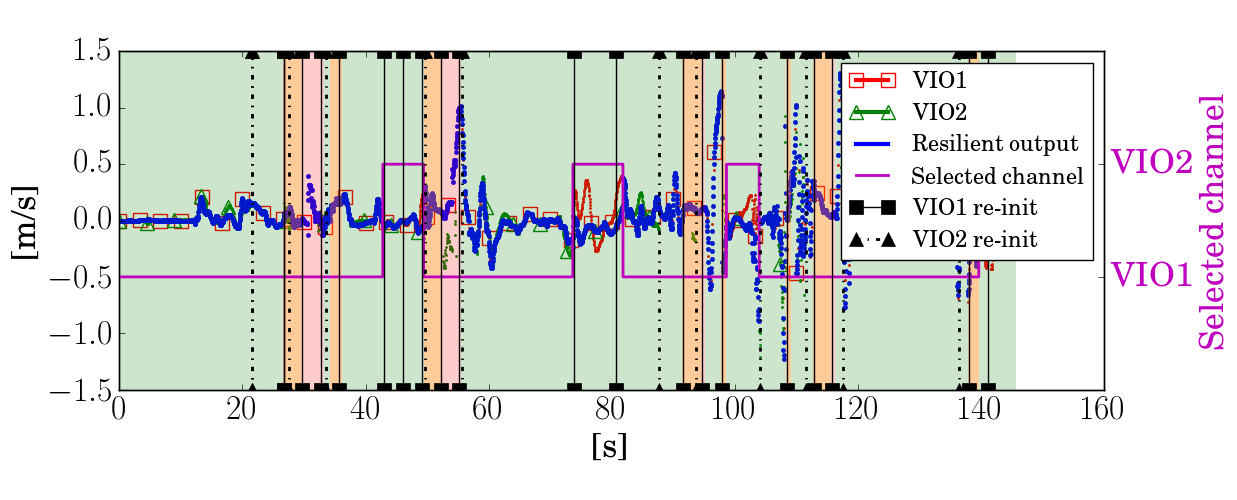} \\
  \rotatebox{90}{\hspace{1.5cm}$v_z$} & \includegraphics[width=0.7\linewidth,trim=0.4cm 1cm 0cm 0.9cm, clip=true]{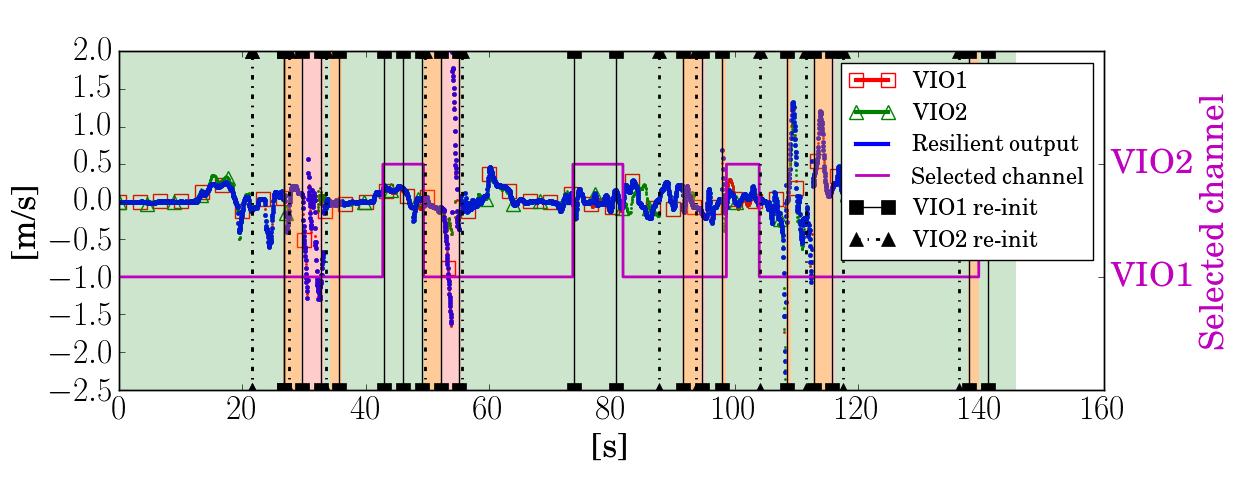} 
\end{tabular}
\vspace{-1em}
\caption{\scriptsize{Results of a real experiment running two VIOs (VIO1 is OS2 and VIO is QSF) together with the resiliency logic. 
In these plots we include: a) Estimations; b) Resiliency logic output (selection of (a) guaranteeing continuity); c) Channel selection to form (b); d) Triggers of re-initialization of methods (vertical lines); and resulting mobility services available depending on the status (coloured areas and right vertical axis).}}
\label{fig:vios_army}
\end{figure}

The modularity of the presented state estimation framework allow us to run different methods on the \ouralgo stack. 
For example, in the case of a drone where the available payload does not allow carring a 3D LiDAR, we can still use redundant VIOs. 
An example of this case is shown in Figure~\ref{fig:vios_army} corresponding to a different experiment in the entrance 2 of the mine. 
In these figures we present the following observations that validates the use of the presented approach:
\begin{itemize}
    \item[\sbt] {\bf VIO estimations:} we show the estimated position (first three plots) and velocities (latter three plots) of OS2 (VIO1 in red with squares) and QSF (VIO2 in green with triangles), together with the respective output of the resiliency logic (blue solid line). Notice how the resilient output does not always overlap with the estimations because it keeps continuity during switches.
    \item[\sbt] {\bf Channel selection:} On all Figure~\ref{fig:vios_army} we overlap the channel selected by the resiliency logic (magenta and specification on the right axis) while switching between methods.
    \item[\sbt] {\bf Re-initialization triggers:} Vertical lines in all plots.
    \item[\sbt] {\bf Mobility services:} Available services depending on the quality of the state estimation, shown in all plots with coloured areas. This includes: Take-off, waypoint navigation and landing using positions (global), wall-following strategies (local) or hovering and landing (attitude).
    In particular for this experiment, for safety reasons we land after 3s of dead-reckoning flight. This strategy helps us to let the dust settle down and thanks to the re-initialization of the methods, we are able to recover the full global mobility services.  
\end{itemize}

\clearpage

\section{Conclusions}\label{sec:conclusions}
In this work, we present a robust and resilient navigation architecture pursuing redundancy ($RP1$), modularity ($RP2$), self-recovery ($RP3$), adaptability ($RP4$) and real-time performance ($RP5$).
This architecture has considerations on the hardware, state estimation and planning and control modules. 

In particular, for hardware we propose to use redundant measurement sources ($RP1$), based on different physical phenomenas to minimize the probability of failures, by mounting sensors in different configurations to maximize the information gain.
Moreover, we advocate for a robust mechanical design of the robot to have significantly higher tolerances to avoid failure.
For state estimation, we present \ouralgo: An heterogeneous and redundant odometry estimator which enables resilient state estimation in perceptually degraded environments.
We run several odometry estimation approaches in parallel($RP1$) that use data from redundant-heterogeneous sensors which are supervised by a resiliency logic. 
\ouralgo is agnostic to the type of odometry approach used and also allows an easy incorporation of commercial-of-the-shelf solutions ($RP2$). The resiliency logic checks the estimation integrity by running confidence tests on sensor data and odometry estimations and, in case of failure, triggers re-initialization of malfunctioning elements ($RP3$). 
Moreover, \ouralgo provides quality estimates which are used by planning and control modules to adapt the mobility behavior of the robot to ensure safety ($RP4$).
In that sense, we use different mobility services depending on the overall state estimation quality, entailing global (\emph{i.e.}, with good global localization), local (\emph{i.e.}, with good velocity estimates) or attitude (\emph{i.e.}, attitude-only estimation) modalities.
Having accurate confidence checks is a critical component to the proposed architecture, an area where further research is welcomed. 
Such research is not only critical to achieving resilient state estimation but can also be in planning behaviors to prevent failures (\emph{e.g.} perception-aware planning).

Our experiments at the DARPA's STIX event of the Subterranean Challenge showed that using the proposed approach we are able to safely navigate in perceptually degraded environments in presence of heavy dust. 
Redundancy ($RP1$) allows the drone to fly for longer distances without failure since it required only one of several estimation sources to work. 
The system is capable of running in real-time ($RP5$) using COTS solutions ($RP2$).
It is able to adapt the behavior ($RP4$) on estimation failures by triggering safety landing and was able to recover ($RP3$) once the dust settled down, resuming the mission.
This architecture, which follows the resiliency principles, enables us to bring the drones in the real-world where failure is not an option.


\section*{Acknowledgement}
This research was carried out at the Jet Propulsion Laboratory, California Institute of Technology, under a contract with the National Aeronautics and Space Administration.

\clearpage

\bibliographystyle{styles/bibtex/splncs_srt}
\bibliography{ISRR_HeRO}

\begin{thebibliography}{10}

\bibitem{Carnevaleijars2015}
Carnevale, D., Martinelli, F.:
\newblock State estimation for robots with complementary redundant sensors.
\newblock International Journal of Advanced Robotic Systems \textbf{12}(10)
  (2015)  138

\bibitem{falanga2018pampc}
Falanga, D., Foehn, P., Lu, P., Scaramuzza, D.:
\newblock {PAMPC}: Perception-aware model predictive control for quadrotors.
\newblock In: IEEE/RSJ International Conference on Intelligent Robots and
  Systems (IROS). (2018)  1--8

\bibitem{MaplessPlanner}
Florence, P., Carter, J., Tedrake, R.:
\newblock Integrated perception and control at high speed: Evaluating collision
  avoidance maneuvers without maps.
\newblock In: Workshop on the Algorithmic Foundations of Robotics (WAFR).
  (2016)

\bibitem{Geiger2012CVPR}
{Geiger}, A., {Lenz}, P., {Urtasun}, R.:
\newblock Are we ready for autonomous driving? the kitti vision benchmark
  suite.
\newblock In: IEEE Conference on Computer Vision and Pattern Recognition.
  (2012)  3354--3361

\bibitem{Gosala2018RedundantPA}
{Gosala}, N., {Bühler}, A., {Prajapat}, M., {Ehmke}, C., {Gupta}, M.,
  {Sivanesan}, R., {Gawel}, A., {Pfeiffer}, M., {Bürki}, M., {Sa}, I.,
  {Dubé}, R., {Siegwart}, R.:
\newblock Redundant perception and state estimation for reliable autonomous
  racing.
\newblock In: IEEE International Conference on Robotics and Automation. (2019)
  6561--6567

\bibitem{kim2019resiliency}
{Kim}, J., {Lee}, C., {Shim}, H., {Eun}, Y., {Seo}, J.H.:
\newblock Detection of sensor attack and resilient state estimation for
  uniformly observable nonlinear systems having redundant sensors.
\newblock IEEE Transactions on Automatic Control \textbf{64}(3) (2019)
  1162--1169

\bibitem{morrell2018comparison}
Morrell, B., Thakker, R., Merewether, G., Reid, R., Rigter, M., Tzanetos, T.,
  Chamitoff, G.:
\newblock Comparison of trajectory optimization algorithms for high-speed
  quadrotor flight near obstacles.
\newblock IEEE Robotics and Automation Letters \textbf{3}(4) (2018)  4399--4406

\bibitem{orbslam2mur2017}
Mur-Artal, R., Tard{\'o}s, J.D.:
\newblock {ORB-SLAM2}: An open-source slam system for monocular, stereo, and
  {RGB-D} cameras.
\newblock IEEE Transactions on Robotics \textbf{33}(5) (2017)  1255--1262

\bibitem{nobili2017heterogeneous}
Nobili, S., Camurri, M., Barasuol, V., Focchi, M., Caldwell, D., Semini, C.,
  Fallon, M.:
\newblock Heterogeneous sensor fusion for accurate state estimation of dynamic
  legged robots.
\newblock In: Robotics: Science and Systems Foundation. (2017)

\bibitem{rollinson2013robust}
Rollinson, D., Choset, H., Tully, S.:
\newblock Robust state estimation with redundant proprioceptive sensors.
\newblock In: ASME 2013 Dynamic Systems and Control Conference. (2013)

\bibitem{Santamaria-NavarroAuro18}
Santamaria-Navarro, A., Loianno, G., Sol{\`a}, J., Kumar, V., Andrade-Cetto,
  J.:
\newblock Autonomous navigation of micro aerial vehicles using high-rate and
  low-cost sensors.
\newblock Autonomous Robots \textbf{42}(6) (2018)  1263--1280

\bibitem{Tomic2012AutonUAV}
{Tomic}, T., {Schmid}, K., {Lutz}, P., {Domel}, A., {Kassecker}, M., {Mair},
  E., {Grixa}, I.L., {Ruess}, F., {Suppa}, M., {Burschka}, D.:
\newblock Toward a fully autonomous {UAV}: Research platform for indoor and
  outdoor urban search and rescue.
\newblock IEEE Robotics Automation Magazine \textbf{19}(3) (2012)  46--56

\bibitem{weissmonocular}
Weiss, S., Achtelik, M.W., Lynen, S., Achtelik, M.C., Kneip, L., Chli, M.,
  Siegwart, R.:
\newblock Monocular vision for long-term micro aerial vehicle state estimation:
  A compendium.
\newblock Journal of Field Robotics \textbf{30}(5) (2013)  803--831

\bibitem{Zhang2015ICRA}
{Zhang}, J., {Singh}, S.:
\newblock Visual-lidar odometry and mapping: low-drift, robust, and fast.
\newblock In: IEEE International Conference on Robotics and Automation. (2015)
  2174--2181

\end{thebibliography}

\end{document}